\documentclass[10pt,twocolumn,letterpaper]{article}

\usepackage{iccv}
\usepackage{times}
\usepackage{epsfig}
\usepackage{graphicx}
\usepackage{amsmath}
\usepackage{amssymb}
\usepackage{cite} 
\usepackage{booktabs}
\usepackage{bbding}
\usepackage{floatrow}
\usepackage{subfigure}
\usepackage{amsmath}
\usepackage{caption}

\usepackage[breaklinks=true,bookmarks=false]{hyperref}

\iccvfinalcopy 

\ificcvfinal\fi

\begin{document}

\title{Learning to Amend Facial Expression Representation via
De-albino and Affinity}

\author{Jiawei Shi, Songhao Zhu$^*$, Dongsheng Wang, Zhiwei Liang\\
College of Automation and Artificial Intelligence\\
 Nanjing University of Posts and Telecommunications, Nanjing, China, 210023\\
{\tt\small \{1319055608, zhush\}@njupt.edu.cn, njuptzsl@yeah.net}
}

\maketitle
\ificcvfinal\thispagestyle{empty}\fi

\begin{abstract}
Facial Expression Recognition (FER) is a classification task that points to face variants. Hence, there are certain affinity features between facial expressions, receiving little attention in the FER literature. Convolution padding, despite helping capture the edge information, causes erosion of the feature map simultaneously. After multi-layer filling convolution, the output feature map named albino feature definitely weakens the representation of the expression. To tackle these challenges, we propose a novel architecture named Amending Representation Module (ARM). ARM is a substitute for the pooling layer. Theoretically, it can be embedded in the back end of any network to deal with the Padding Erosion. ARM efficiently enhances facial expression representation from two different directions: 1) reducing the weight of eroded features to offset the side effect of padding, and 2) decomposing facial features to simplify representation learning. Experiments on public benchmarks prove that our ARM boosts the performance of FER remarkably. The validation accuracies are respectively \textbf{90.42\%} on RAF-DB, \textbf{65.2\%} on Affect-Net, and \textbf{58.71\%} on SFEW, exceeding current state-of-the-art methods. Our implementation and trained models are available at \href{https://github.com/JiaweiShiCV/Amend-Representation-Module}{https://github.com/JiaweiShiCV/Amend-Representation-Module}.
\end{abstract}

\textbf{Index Terms}: Facial Expression Recognition, Image Classification, Representation Learning

\section{Introduction}

	Facial expression is the external reflection of emotion, and in essence, the movement of subcutaneous muscles. In interpersonal communication, the facial expression is one of the most important media for people to convey feelings and attitudes to each other. Therefore, facial expression recognition is of great significance for human-computer interaction. 

	With GPU performance growing drastically, deep learning has come a long way in just over a decade. Gradually, deep learning has become the mainstream method to promote FER. Relying on the famous neural networks in the classification era, some scholars put forward their algorithms to deal with the challenges of FER. Wang \etal \cite{scn} propose Self-Cure Network based on ResNet-18 \cite{resnet} to suppress the uncertainties of data annotation in network training.

\begin{figure}[t]

\centering
	\subfigure[]{\includegraphics[width=0.4\linewidth]{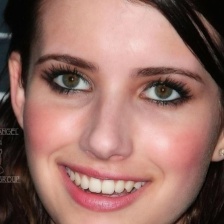}
		\label{Fig. 1(a)}}\hspace{10mm}
	\subfigure[]{\includegraphics[width=0.4\linewidth]{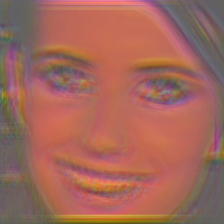}
		\label{Fig. 2(b)}}
	\caption{(a) An original facial expression image, and (b) its appearance after padding convolution. The edges of the expression are eroded by the padding.} 
\label{fig:padding}

%
\end{figure}

	The well-known neural networks applied to FER have achieved certain effects, whereas they cannot fully give play to the learning advantages of the deep network because of the limitation to the scale of the FER dataset. Creating larger datasets brings prohibitive annotation costs. How to improve the recognition accuracy from the limited datasets has become a major challenge for FER.

	Not only that, but we find that certain intrinsic properties of the convolutional neural network limit and hinder its ability. To encode information accurately on the image, the edge information cannot be ignored. However, the edges of the image, especially the corners, are perceived less than the inside pixels by the convolution kernel. The way the convolutional kernel works makes the weight of the information on the edges lower than that in the middle. We call the negligent property of convolutional kernel as perception bias. Therefore, it is accustomed to utilize padding, a simple process of adding zeros to the periphery of the images, to balance the perceptual frequency of each pixel. The same is true for the feature map. When the network is fairly shallow, it seems unnecessary to consider the influence of a few zeros on the formation of features, because the image pixels are relatively huge. Normally, as the network deepens, the channels increase exponentially. Conversely, the area of the feature map keeps shrinking. The pixel information becomes scarce and refined. That makes the deep convolutional layers show less and less receptive to the fade of the edges, as edge pixels account for a considerable part of the total. The only countermeasure is to keep padding to ensure the capture of key information. Since then, the network has fallen into the \emph{padding trap}. The essence of convolution is to calculate the sum of the products of the pixel values in each region and those of the kernel. Excessive zeros involved will inevitably lead to serious information distortion. This is a serious issue that we have been neglecting in the past. As shown in Figure \ref{fig:padding}, just one convolution operation with padding, the periphery of the image becomes blurred visible to the naked eye. Layer by layer, padding gradually erodes the feature map from outside to inside. We call the information that gravely impaired in this way albino features. Actually, the unprocessed albino features directly affect the representations downward.

	Hence, in order to avoid the problems mentioned above, we need to consider not only the internal correlation hide in the FER datasets but also how to eliminate the erosion of representation. In this work, we propose a module that can be embedded in any network with a pooling layer, called Amend Representation Module (ARM), which consists of an auxiliary block and two functional blocks. 

\begin{figure}
\begin{center}
\includegraphics[width=0.98\linewidth]{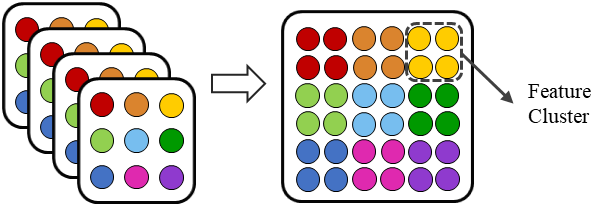}
\end{center}
   \caption{A case to demonstrate our rearrangement method for feature map. This process only changes the absolute position of the feature points, and the relative position remains. The channel collapses in a ratio of 4. We name the gathered points which are originally at the same position as a feature cluster.}
\label{fig:arrangement}
\end{figure}

	To protect representation from erosion, we remove the pooling layer and replace it with De-albino (DA) block. The Feature Arrangement (FA) block, customized to assist the DA block to work better, arranges the raw feature map into a single channel under the premise of minimum position change of feature pixels. As shown in Figure \ref{fig:arrangement}, all the pixels that are severely eroded are concentrated in the outermost part of the feature map. Then we take advantage of the perception bias of the convolutional kernel to subtly reduce the weight of the albino features. The perception bias is positively correlated with the size of the convolutional layer kernel. It is worth noting that padding is resolutely denied throughout the module. Because we regard it as the culprit of the problem. 

	Secondly, inspired by the nature of facial expressions, which are variants of the human face, we consider that there are certain affinities among expression features. We explore the role of this relationship and obtain a benign response of FER performance. Sharing Affinity (SA) block treats the face feature as a combination of a generic part and a unique part, and simplifies the representation learning by splitting the features, which improves the feature learning of the classic network on the FER dataset.  

	Most of the previous FER strategies are to extend the network structure or supplement the algorithm on the classic CNNs. While our ARM is an internal modification of the Convolutional Neural Network (CNN). By training the ARM variant of ResNet-18 \cite{resnet} on RAF-DB, we promote the performance of pre-trained ResNet-18 from 86.4 to 90.42, which is superior to the-state-of-art methods.

Our contribution is mainly divided into the following parts:

(1) We put forward for the first time the Padding Challenge, which is to eliminate the albino erosion existing in the padding convolution.

(2) We propose a de-albino strategy, which serves all the convolutional neural networks, to protect representations from albino features.

(3) We elaborately design an auxiliary block to rearrange the feature map, enhancing the de-albino effect.

(4) We make the first attempt to optimize the FER with affinity features among different facial expressions.

(5) We design a minimal random resampling scheme to deal with data imbalance.

\section{Related Work}
\subsection{Convolutional Neural Networks}

	In the late 1990s, Yann LeCun \etal \cite{lecun}, in order to solve the task of handwritten digit recognition, proposed LeNet, in which he first applied BP algorithm to the training of neural network structure. This feat established the modern structure of CNN: convolutional layer, pooling layer, and fully connected layer. However, ascribe to the lack of data and computing power, LeNet is difficult to deal with complex problems. Despite the effectiveness of identifying numbers, LeNet is not as good as machine learning algorithms in practical tasks.

	Gradually, with the rapid progress of the GPU unit and the completion of large-scale datasets, CNN came back to the attention of researchers. In the 2012 ImageNet Image Recognition Contest, AlexNet, a brand new structure introduced by Hinton \etal \cite{alexnet}, was the winner by an absolute advantage of 10.9 percentage points over the runner-up. It came to use padding to balance the weights of the pixels between the edges and the inside. Instead of Sigmoid, they applied the ReLU \cite{relu} activation function to solve the gradient dispersion of the deep network. The proposed Dropout and Data Augmentation methods \cite{alexnet} are effective in preventing overfitting. It was the first time that people realized that the structure of neural networks had a lot of room for improvement. In the years that followed, various CNNs sprang up.

	In the 2014 ILSVRC Competition, GoogLeNet \cite{googlenet}, the winner, was characterized by the application of the Network in Network structure, that is, the original node is also a Network. Karen Simonyan and Andrew Zisserman \cite{vgg} who proposed the VGG architecture won second place. They tended to use smaller kernels to deepen the network. This attempt effectively improved the performance of the model. It was also proved that VGG had a good ability to generalize on other datasets and outperformed GoogLeNet in multiple transfer learning tasks. In 2015, He Kaiming \etal \cite{resnet} from MSRA solved the degradation problem of the extremely deep network through the cross-layer transmission of Identity. They deepened the network to hundreds or even thousands of layers. The error rate also dropped to 3.6\%, lower than the human rate of 5.1\%. In the last ILSVRC competition, Hu J \etal \cite{senet} proposed SENet, which recalibrates the channel dimension by squeezing features in the spatial dimension and extracting the correlation between channels, thus further reducing the error rate.

\subsection{Facial Expression Recognition}

	Facial Expression Recognition is a classification task of identifying the facial changes in response to human emotions from visual information. This technology enables machines to understand human emotions and make judgments in response to mental states. The research on facial expressions developed earlier in psychology and medicine. As early as the 19th century, Charles Darwin and Phillip Prodger \cite{darwin} explained the difference and connection of facial expressions between humans and animals. In 1971, Ekman and Friesen \cite{ekman} defined the six basic human expressions (that is \emph{happiness}, \emph{sadness}, \emph{surprise}, \emph{fear}, \emph{anger}, and \emph{disgust}), then systematically established a facial expression image library, which describes the details of each expression. In 1978, Suwa \etal \cite{suwa} proposed automatic expression analysis in facial video. However, due to the limitations of contemporary computer capabilities, facial expression research in the computer field started relatively late. In the 1990s, Kenji Mase \cite{mase} proposed to leverage the optical flow method for automatic expression analysis, opening the computer era of facial expression recognition. In 2001, Tian \etal \cite{tian} utilized action units (AU) instead of the traditional expression images to classify facial expressions, creating a precedent for automatic expression analysis. 

	Early feature learning methods (\eg, non-negative matrix factorization (NMF) \cite{nmf}, local binary patterns (LBP) \cite{lbp}, histograms of oriented gradients (HOG) \cite{hog}, scaled-invariant feature transform (SIFT) \cite{sift}, etc.) are costly and inefficient, remaining at the level of manual annotation and calculation. As datasets expand into the wild, such as FER2013 \cite{fer2013}, AffectNet \cite{affectnet}, EmotioNet \cite{emotionet}, SFEW \cite{sfew}, and RAF-DB \cite{raf-db}, the engineered learning methods are difficult to cope with data extremely large-scale and complex. With the rapid development of GPU units, such algorithms have gradually faded and been replaced by deep learning. Not only learning deep features, but convolutional neural networks can also dynamically adjust neuron weights based on the deviation between the prediction and the target. Heechul Jung \etal \cite{jung} design an integration network to process facial images and landmark points jointly. Huiyuan Yang \etal \cite{yang} believe that expression is a superposition of neutral components and expression components. They exploit a network that generates neutral expressions to isolate residual expression components. Feifei Zhang \etal \cite{zhangfeifei} combine different poses and expressions to train GAN for pose-invariant expression recognition. Kai Wang \etal \cite{scn} propose the Self-Cure algorithm to relabel uncertain expressions, which inhibited the network from overfitting incorrectly labeled samples. In order to recognize the occluded expressions, Bowen Pan \etal \cite{pan} apply a network pre-trained on an unoccluded data set to guide the fine-tuning of the occluded network. To solve the problem of label inconsistency when multiple datasets are merged, Jiabei Zeng \etal \cite{zeng} generate multiple pseudo-labels for images and then train the model in this environment to learn the underlying truth. Shikai Chen \etal \cite{chenshikai} generate multi-labels for label distribution learning by classifying action units and facial landmarks separately.

\begin{figure*}
\begin{center}
\includegraphics[width=0.98\linewidth]{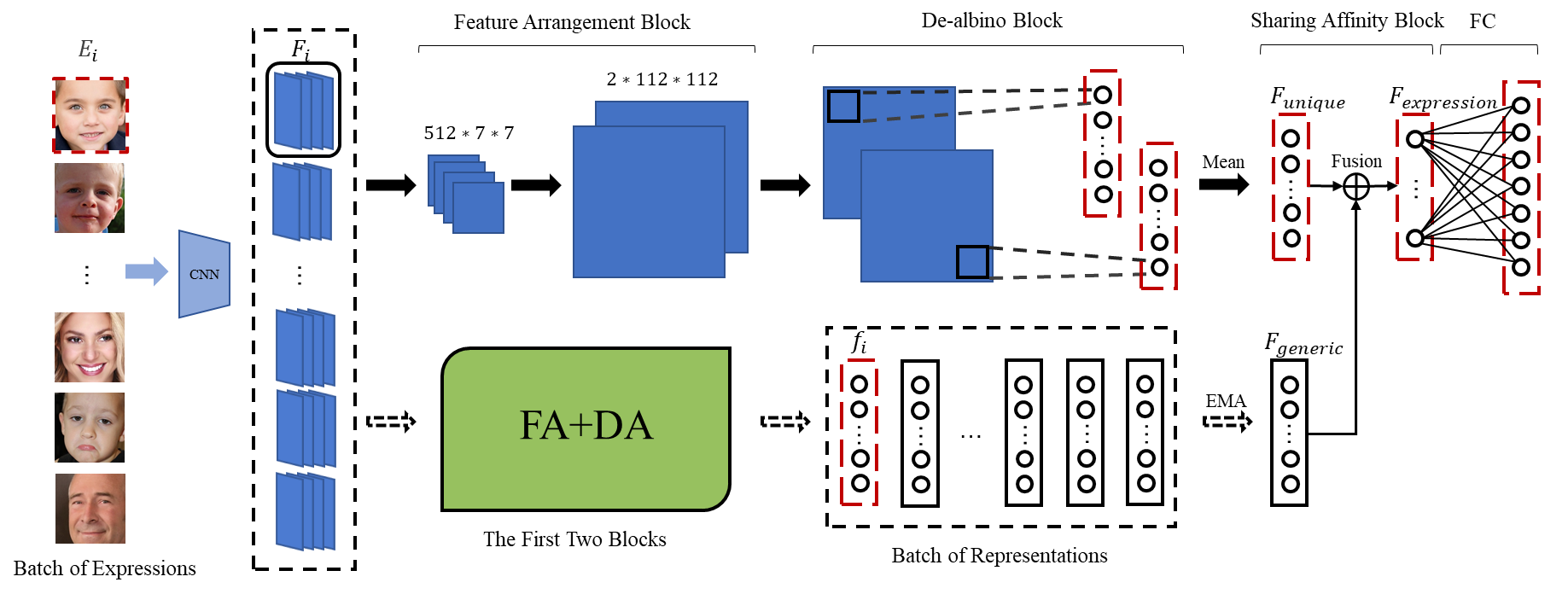}
\end{center}
   \caption{Overview of Amend Representation Module (ARM). The ARM composed of three blocks replaces the pooling layer of CNN. The solid arrows indicate the processing flow of one feature map, and the dotted arrows refer to the auxiliary flow of a batch. It should be noted that the relationship between the two channels requires the de-albino kernel to be single-channel and unique.}
\label{fig:Net}
\end{figure*}

\section{Methodology}
	The padding is crucial to convolutional layers because of functionality. Although boosting the performance of the network \cite{lecun, alexnet}, it may be harmful to the representation because of the introduction of a large amount of information, which is foreign to the image. Sticking to the regular pooling layer does nothing about the padding's negative impacts. Conjunctively, considering the particularity of the Facial Expression Recognition (FER) task, that is, the affinity features between facial expressions, we propose the Amend Representation Module (ARM). In this section, we overview the ARM and then describe the function of each component in detail.
\subsection{Overview}
	Given a dataset of facial expressions, firstly, we extract the feature of each image with the front part of the ResNet-18 \cite{resnet}, which is bounded by Global Average Pooling (GAP). Then, the learned feature map is fed into our ARM, which outputs high-quality representations.

	As shown in Figure \ref{fig:Net}, our ARM consists of three crucial blocks, namely Feature Arrangement (FA) block, De-albino (DA) block, and Sharing Affinity (SA) block. The FA block is an auxiliary block to amplify the function of the DA block. The latter realizes the weight distribution of features by means of convolution. Naturally, the weight with a higher degree of albino erosion is lower, and vice versa. The final SA block splits facial features into two parts for simple and efficient learning.

\subsection{Feature Arrangement}\label{section:arrangement}
Due to convolutional padding, the formation of albino features is inevitable. With reference to the position of the padding, it can be inferred that the edges and corners are where the de-albino features gather. The albino pixels are evenly dispersed in each channel, which is not conducive to “targeted therapy” because a slight expansion of the de-albino range will cause damages to key information. We try to find a way to bring the same degree of albino pixels together, but maintain their relative position distribution. That is to keep the most eroded pixels concentrated on the periphery of the feature map. The method mentioned in the Effective Sub-Pixel Convolution layer \cite{sub} meets the demand, so we adopt it to restructure the feature map. The specific implementation is shown in Figure \ref{fig:arrangement}. Then a no-padding block is used to weaken the weight of the albino information at the edge and neutralize the baneful influence of padding.
\begin{figure}
\begin{center}
\includegraphics[width=0.5\linewidth]{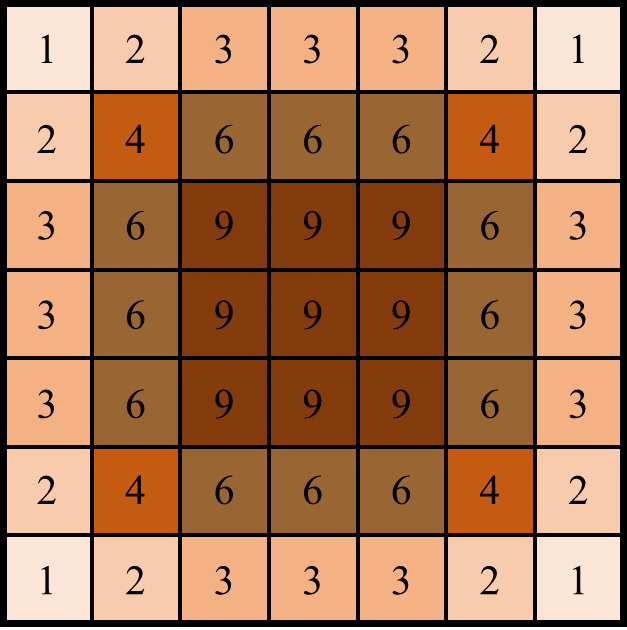}
\end{center}
   \caption{Illustration of the perception bias of the no-padding convolution on 7x7 pixels. The convolution kernel size and stride are 3 and 1, respectively. The numbers on the pixels indicate the perceived frequency.}
\label{fig:perception}
\end{figure}

\subsection{Assigning Weights by Means of Convolution}
To eliminate the impact of albino features, we elaborately designed a dedicated convolutional layer to dilute the weight of the peripheral pixels. It is characterized by no padding, large convolutional kernel, and big stride.

As shown in Figure \ref{fig:arrangement}, the most eroded points are concentrated at the edge of the rearranged feature map. We subtly exploit the perception bias illustrated in Figure \ref{fig:perception} to reduce the weights of albino points. The pixels at the edge are perceived less frequently by the convolutional kernel than the internal ones. Considering the absolutely large scale of the concentrated albino points, a regular-sized convention kernel (such as 3x3) is not suitable for our method. And it can be concluded that the larger the convolutional kernel, the more information is ignored, which happens to favor the network in this case. Because of abundant channel information flowing into space, the convolutional kernel that we take should be several times larger than the regular one. The stride of the kernel also has an effect on the perception of peripheral points. We filter the albino features and retain the key parts by adjusting the size and stride of the convolution kernel.

\subsection{Sharing Affinity}
Convolutional neural networks tend to assign weights to features favoring evaluation indicators, which makes the learning short-sighted. Different from other classification tasks, Facial Expression Recognition is a task based on face variants. Yang \etal \cite{yang2018} consider facial expression as a combination of an expressive component and a neutral component. Since the face is the main carrier, there are affinity features between the facial expressions. 

	We regard facial expressions as two parts: universal and unique:

\begin{equation}
F_{expression} = F_{generic} + F_{unique} 
\end{equation}

 With respect to similar human face, learning the generic feature can simplify the feature learning. That means the network does not need to express the complete features of faces from scratch, but only learn differences based on generic features. Given a batch of facial expressions, we extract the average representations as initial generic representation, and then update it using Exponential Moving Average (EMA):

\begin{equation}
F_{batch}, F_{init} = \frac{\sum_{i=1}^{N}f_i}{N} 
\end{equation}


\begin{equation}
F_{generic}=\lambda F_{batch}+(1-\lambda) F_{generic}^{'}
\end{equation}
The $F_{batch}$ and $F_{generic}^{'}$ denote the average feature of the current batch and the previous $F_{generic}$ respectively. The coefficient $\lambda$  represents the degree of weighting decrease, a constant smoothing factor between 0 and 1. A higher $\lambda$ discounts older observations faster, which can be either a fixed value or a learnable parameter.

\begin{figure*}[t]
\begin{center}

\centering
	\subfigure[]{\includegraphics[width=0.5\linewidth]{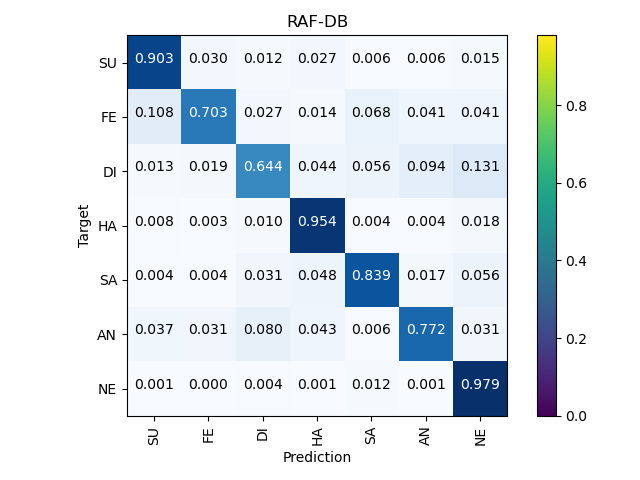}
		\label{Fig. 1(a)}}\hspace{-5mm}
	\subfigure[]{\includegraphics[width=0.5\linewidth]{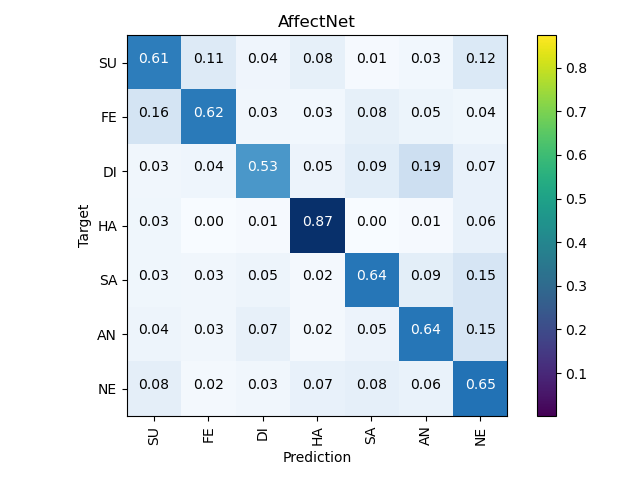}
		\label{Fig. 2(b)}}
	\subfigure[]{\includegraphics[width=0.5\linewidth]{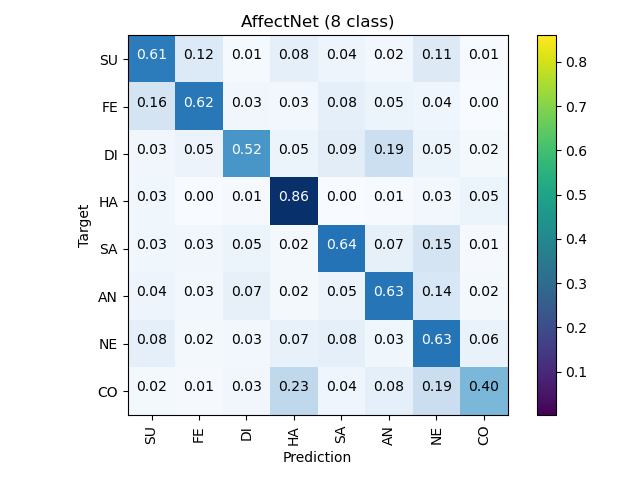}
		\label{Fig. 3(c)}}\hspace{-5mm}
	\subfigure[]{\includegraphics[width=0.5\linewidth]{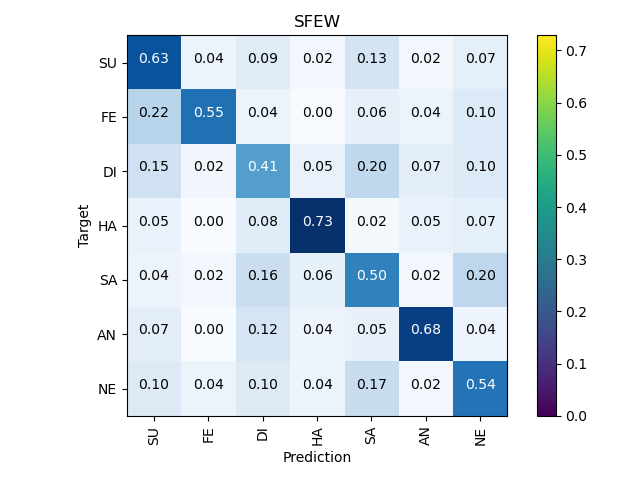}
		\label{Fig. 4(d)}}
	\caption{The confusion matrixes on the test set for the RAF-DB, AffectNet (7cls \& 8cls), and SFEW.}
\label{fig:fusion}

\end{center}
\end{figure*}

\begin{table*}[t]
\begin{center}
\begin{tabular}{cccccccccc}
\toprule
	Category & Neutral & Happy & Sad & Surprise & Fear & Disgust & Anger & Contempt & Total\\
	\midrule
	images & 74,874 & 134,415& 25,459& 14,090& 6,378& 3,803& 24,882& 3,750& \textbf{287,651} \\
\bottomrule
\end{tabular}
\end{center}
\caption{The statistics of manually annotated images in the AffectNet training set.}
\label{tab:affectnet}
\end{table*}

\section{Experiments}
	In this section, the ARM, proposed deep learning architecture, is evaluated on three FER datasets. We first explain why we choose these datasets and what they are. A description of metrics followed closely. We conduct a verification experiment to demonstrate the adverse effect of albino features. Then we introduce the overall details of the experiment and indicate the fine-tuning on the AffectNet dataset. To evaluate each block of the ARM, an ablation study is performed on RAF-DB dataset. Finally, our ARM is compared with the state-of-the-art methods.

\subsection{Datasets}
	To evaluate our module, we select three public FER datasets collected in the wild that fully reflect real-world emotions. Compared with those produced in the laboratory, there is no deliberate pose before generating. They are more complex, variable, and challenging for feature learning. The details are as follows.

	\textbf{RAF-DB} \cite{raf-db} is a crowdsourced dataset. Each image was individually labeled by 40 taggers for the sake of rigor. There are two distinct subsets, the basic one being single-tagged and the compound one being double-tagged. Only the former is used in our experiment. The basic dataset contains seven categories: surprise, fear, disgust, happiness, sadness, anger, and neutral. The number of the training set and testing set is 12,271 and 3,068 respectively, and the expressions of them have near-identical distribution. 

	\textbf{AffectNet} \cite{affectnet} is a large facial expression dataset, with a total of more than one million images. The main collection methods are searching through three search engines and downloading from the Internet. Images are annotated in two ways: manual and machine. Additionally, it contains ten categories, eight of which are basic expressions. In the manually labeled subset, excluding the extra \emph{contempt}, the same seven basic expressions as in RAF-DB are used, namely 283,901 training images and 3,500 validation images. We also conduct eight types of basic expression experiments for better comparison. Because of the lack of a testing set, we fill this void with the validation set. Remarkably, while the categories are balanced on the validation set, they are extremely unbalanced on the training set, with a maximum gap of over 35 times. 

	\textbf{SFEW} \cite{sfew} contains 958 training samples, 436 validation samples and 372 test samples, each of which is the static frame selected from Acted Facial Expressions in the Wild (AFEW) \cite{afew} dataset. These facial images also fall into seven basic categories and have close distribution in three sets.

\subsection{Metrics and Results}
	The overall sample accuracy is the most common and straightforward metric, namely accuracy or weighted accuracy \cite{psr}. In simple terms, it is calculated by dividing the number of correctly predicted samples by that of testing samples. Unlike the former, mean class accuracy, which is also called unweighted accuracy \cite{psr}, is the average of the accuracy of each category. The sum of the accuracy of each class is first calculated and then divided by the number of classes. 

	Ascribe to the learning bias of CNN, categories that are easy to distinguish or have a large number of samples will gain more weight in neurons \cite{imbalance}. Under the circumstances, the overall accuracy determined by a biased category may still be at a high level, whereas its performance is certainly not as good as the accuracy appears. To prevent overall sample accuracy from masking the illusion of good performance, we introduced unweighted accuracy as a backup metric.

	Figure \ref{fig:fusion} shows the confusion matrixes of the ARM (ResNet-18) on RAF-DB, AffectNet (7cls \& 8cls), and SFEW. It's rarely seen in other strategies that \emph{neutral} hits the highest accuracy of 98\% on RAF-DB, slightly exceeding the \emph{happiness}.
\begin{figure}[t]
\begin{center}
\includegraphics[width=0.9\linewidth]{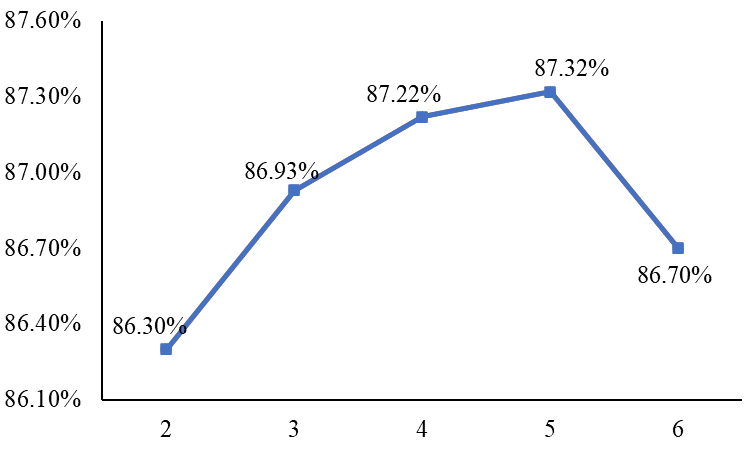}
\end{center}
   \caption{Evaluation of the kernel size $k$ of the De-albino block on RAF-DB dataset.}
\label{fig:albino}
\end{figure}

\subsection{The Adverse Effect of Albino Features}

	For the sake of scientific rigor, we conduct an independent experiment to demonstrate the negative effect of albino features. For the network part, ResNet-18 \cite{resnet} was modified to meet experimental requirements. Specific implementation details are as follows: De-albino block is inserted to replace the GAP to carry out convolution processing on the output 512x7x7 feature map. In the De-albino block, the kernel size is denoted as $k$, adjustable and the stride is fixed as 1. Finally, keep the fully connected layer and make it adaptive. 

	The experimental results are shown in Figure \ref{fig:albino}. Viewpoints can be analyzed from the figure. With the increase of $k$, the gain effect presents a hump-like trend which fits with the fluctuation of the perception frequency ratio between the center and the edge. Experiments on RAF-DB show beyond doubt that, by reducing the weight of albino features on the periphery of the feature map, the quality of representation can be effectively improved, thus slightly boosting the performance of the model. On the other hand, albino features do erode the representation of images.
\begin{table}[t]
\begin{center}\resizebox{\textwidth}{!}{
\begin{tabular}{cccc}
\toprule
	Network & Output & Hyperparameter & Param num\\
	\midrule
ResNet-18$^*$ & [None, 512, 7, 7]  & / & 11,178,395\\
Arrangement & [None, 2, 112, 112] & - & \textbf{0}\\
De-albino & [None, 2, 11, 11] &c=1, k=32, s=8, p=0 & \textbf{1024}\\
BN &    [None, 2, 11, 11]    & -  & \textbf{4}\\  
Mean & [None, 11, 11] & - & \textbf{0}\\
Affinity & [None, 11, 11] & Parameter $\lambda$  & \textbf{1}\\
FC & [None, 7] & 121-7 & 854\\
\bottomrule
\end{tabular}}
\end{center}
\caption{Network with ARM for RAF-DB. ResNet-18$^*$ retains the part before GAP.}
\label{tab:network}
\end{table}

\begin{table*}[!tp]
\caption{Comparison with the state-of-the-art methods on different datasets. Two metrics are used on RAF-DB, namely Weighted Accuracy (WA) and Unweighted Accuracy (UA).}
\centering
\subtable[Comparison on RAF-DB.]{\resizebox{0.315\textwidth}{!}{
       \begin{tabular}{ccc}
    \toprule
    Method & Acc. & mean Acc.\\
    \midrule
    DLP-CNN \cite{raf-db} & 84.22  & -   \\
    IPA2LT \cite{zeng}  & 86.77  & -  \\
	RAN \cite{ran} &  86.90  & -     \\
	SCN \cite{scn}& 87.03 & - \\
	DACL \cite{dacl}& 87.78 & 80.44 \\
	PSR \cite{psr}& 88.98 & 80.78  \\
    \midrule
	ResNet-18 \cite{resnet} & 84.68 & 77.63	\\
    ResNet-18 (ARM)     & \textbf{90.42} & \textbf{82.77} \\
    \bottomrule
\end{tabular}}
}
\hspace{-0.1cm}
\subtable[Comparison on AffectNet.]{   \resizebox{0.365\textwidth}{!}{     
       \begin{tabular}{ccc}
    \toprule
    Method & Acc. (7 cls) & Acc. (8 cls)\\
    \midrule
    Upsample \cite{affectnet} & - & 47.00      \\
    IPA2LT  \cite{zeng} & - & 57.31    \\
	RAN \cite{ran}&  - & 59.50       \\
	LLHF \cite{llhf} & 63.31 & 59.58 \\
	SCN  \cite{scn}& - & 60.23 \\
	PSR \cite{psr} & 63.77 & 60.68  \\
    \midrule
	ResNet-18 \cite{resnet} & 51.56 & 47.61	\\
    ResNet-18 (ARM)    & \textbf{65.2} & \textbf{61.33} \\
    \bottomrule
\end{tabular}}
}
\hspace{-0.1cm}
\subtable[Comparison on SFEW.]{   \resizebox{0.245\textwidth}{!}{     
       \begin{tabular}{cc}
    \toprule
    Method & Acc. \\
    \midrule
	Identity-aware CNN \cite{identity} & 50.98 \\
	Island Loss \cite{island}	 & 52.52 \\
	Multiple deep CNNs \cite{multiple} & 55.96 \\
	RAN \cite{ran}   & 54.19\\

	\midrule
	ResNet-18 \cite{resnet} & 51.56	\\
	ResNet-18 (ARM) & \textbf{58.71} \\
    \bottomrule
\end{tabular}}
}
\label{tab:comparison}
\end{table*}

\subsection{Implementation Details}
	The backbone network is part of ResNet-18, where the layers before the GAP are retained and initialized with weights pre-trained on ImageNet \cite{imagenet}. We leverage this part of the network to extract the albino features, which are in the size of 512x7x7. 

	 Based on the rearrangement principle described in Section \ref{section:arrangement}, only the feature map of two channels can be obtained because the channel number is not an integer multiple of 4. As the channels of the feature map decrease, the spatial area of the feature map increases to 112x112 pixels. It is essential to emphasize that the feature map does not ideally lie flat in a single channel, whereas its channels should still be considered parallel. Considering the correlation of the two channels, the kernel of the De-albino block must be single-channel. The point set composed of features that are from the same position of each channel is called \emph{feature} \emph{clusters}. It's in the size of 16x16x1. 

	First, there is no doubt that padding will not be in our ARM. Then a proper convolutional kernel is definitely crucial to the experimental results. Obviously, if the size of the convolution kernel is slightly larger than that of the \emph{feature cluster}, sometimes a \emph{feature cluster} plays an absolute role within the sphere of convolution kernel, which weakens the diversity of the representation. So we set the convolutional kernel as 32x32x1 pixels, much larger than the \emph{feature cluster}. Meanwhile, the stride of the convolutional kernel is set to half of the cluster size, which is 8. The two channels output from the DA block are averaged as the representation of each expression.

	As for the SA block, the gain hyperparameter $\lambda$ varies depending on the affinity within the dataset. By default, it is initialized to 0.3 and updated with the gradient.

	The details of the network are listed in Table \ref{tab:network}. According to the outputs of the Network, we can efficiently obtain high-quality representations from the variant of ResNet-18. To optimize the parameters of the network, we adopt the Adam solver \cite{adam} with mini-batch size 256 on RAF-DB. The learning rate is initialized as 0.001, and we apply a decay learning rate strategy with a coefficient of 0.9.

The Amend Representation Module (ARM) is implemented using the PyTorch toolbox \cite{pytorch}. All of the reported results are obtained by running the Python code on an NVIDIA GeForce RTX 3090 GPU.

\begin{table}[t]
\begin{center}\resizebox{\textwidth}{!}{
\begin{tabular}{cccc}
\toprule
	Arrangement & De-albino & Affinity & RAF-DB(pretrained)\\
	\midrule
$\times$ & $\times$ & $\times$ & 84.68\\
$\times$ & $\checkmark$ & $\times$ & 87.32\\
$\times$ & $\times$ & $\checkmark$ & 87.03\\
$\checkmark$ & $\checkmark$ & $\times$ & 88.54\\
$\checkmark$ & $\times$ & $\checkmark$ & 87.40\\
$\times$ & $\checkmark$ & $\checkmark$ & 87.38\\
$\checkmark$ & $\checkmark$ & $\checkmark$ & 90.42\\
\bottomrule
\end{tabular}}
\end{center}
\caption{Evaluation of the three blocks in ARM.}
\label{tab:ablation}
\end{table}

\subsection{Fine-tuning on the AffectNet Dataset}
	As shown in Table \ref{tab:affectnet}, the sample numbers in each category in AffectNet \cite{affectnet} is extremely unbalanced. Intuitively speaking, in the 8 basic categories, the ratio of the largest to smallest sample number is as high as 35. 

	For superior performance, we prudently designed a minimal random resampling (MRR) strategy to balance the numbers of categories. The implementation details are as follows: First, we assign a probability of selection to each expression, which is the reciprocal of the sample size. In this case, each image has the same probability of being picked after a random selection. During training, the sample size of each category is uniformly set to be consistent with the minimum category. Although the size of our training set has shrunk greatly from the original, we retain the scale advantage of the original dataset beacause the training set is different for each epoch. After multiple epochs of training, even the largest category can basically be traversed. It should be noted that we did not discard the massive amount of original data like Undersampling because the training set of each epoch is randomly re-sampled. The Oversampling strategy is not adopted because Oversampling will duplicate the minimum category dozens of times, which inevitably leads to serious overfitting of the model. The samples inside the superclasses may be completely different in different epochs of training. After all, these classes are only sampled at a minimum of one-35th, so we reduced the learning rate by adjusting its attenuation rate from 0.9 to 0.78.

\subsection{Ablation Study}
	The effectiveness of three blocks in ARM. In order to better understand the respective function of each block, we conducted an ablation study to evaluate them on RAF-DB. We show the experiment results in Table \ref{tab:ablation}. According to the original intention of the design, the Feature Arrangement block is auxiliary, so we did not conduct a separate experimental evaluation for it. Analyzing the data, we can draw the following points. First of all, the main functional blocks, namely the DA and SA blocks, have obvious effects, and they can improve the network performance separately. Secondly, the DA block and the SA block react differently with the FA block. With the assistance of the FA block, the DA block can increase performance by 1\% again, but the SA block does not seem to benefit from it. Observing the complete coordination performance of the three blocks, it can be seen that the SA block performs significantly on high-quality representation purified by the FA block and DA block. Therefore, it is better to use the SA block when the current representation is sufficiently accurate. 

\subsection{Method Comparison}
	In Table \ref{tab:comparison}, we compare the proposed method with a series of state-of-the-art baselines, which are briefly described below.
	Our ARM outperforms current state-of-the-art methods by a large margin, with \textbf{90.42\%}, \textbf{65.2\%}, and \textbf{58.71\%} on RAF-DB, AffectNet, and SFEW, respectively.
\section{Conclusion}

We develop a lightweight module to solve the potential Padding Erosion that has always been neglected. The role is amplified through the auxiliary module, showing a significant improvement in the facial expression recognition task.  

\noindent\textbf{Acknowledgments} 

This work is supported by Natural Science Foundation of Nanjing University of Posts and Telecommunications under No. NY219107, and National Natural Science Foundation of China under No. 52170001.



{\small
\bibliographystyle{ieee_fullname}
\bibliography{egbib}

\begin{thebibliography}{10}\itemsep=-1pt

\bibitem{imbalance}
Mateusz Buda, Atsuto Maki, and Maciej~A Mazurowski.
\newblock A systematic study of the class imbalance problem in convolutional
  neural networks.
\newblock {\em Neural Networks}, 106:249--259, 2018.

\bibitem{fer2013}
P.~L. Carrier, A. Courville, I.~J. Goodfellow, M. Mirza, and Y. Bengio.
\newblock {\em FER-2013 face database.}
\newblock Universit de Montreal, 2013.

\bibitem{chenshikai}
Shikai Chen, Jianfeng Wang, Yuedong Chen, Zhongchao Shi, Xin Geng, and Yong
  Rui.
\newblock Label distribution learning on auxiliary label space graphs for
  facial expression recognition.
\newblock In {\em Proceedings of the IEEE/CVF Conference on Computer Vision and
  Pattern Recognition}, pages 13984--13993, 2020.

\bibitem{hog}
Navneet Dalal and Bill Triggs.
\newblock Histograms of oriented gradients for human detection.
\newblock In {\em 2005 IEEE computer society conference on computer vision and
  pattern recognition (CVPR'05)}, volume~1, pages 886--893. Ieee, 2005.

\bibitem{darwin}
Charles Darwin and Phillip Prodger.
\newblock {\em The expression of the emotions in man and animals}.
\newblock Oxford University Press, USA, 1998.

\bibitem{imagenet}
Jia Deng, Wei Dong, Richard Socher, Li-Jia Li, Kai Li, and Li Fei-Fei.
\newblock Imagenet: A large-scale hierarchical image database.
\newblock In {\em 2009 IEEE conference on computer vision and pattern
  recognition}, pages 248--255. Ieee, 2009.

\bibitem{sfew2.0}
Abhinav Dhall, Roland Goecke, Simon Lucey, and Tom Gedeon.
\newblock Static facial expression analysis in tough conditions: Data,
  evaluation protocol and benchmark.
\newblock In {\em 2011 IEEE International Conference on Computer Vision
  Workshops (ICCV Workshops)}, pages 2106--2112. IEEE, 2011.

\bibitem{ekman}
Paul Ekman and Wallace~V Friesen.
\newblock Constants across cultures in the face and emotion.
\newblock {\em Journal of personality and social psychology}, 17(2):124, 1971.

\bibitem{emotionet}
C Fabian Benitez-Quiroz, Ramprakash Srinivasan, and Aleix~M Martinez.
\newblock Emotionet: An accurate, real-time algorithm for the automatic
  annotation of a million facial expressions in the wild.
\newblock In {\em Proceedings of the IEEE conference on computer vision and
  pattern recognition}, pages 5562--5570, 2016.

\bibitem{dacl}
Amir~Hossein Farzaneh and Xiaojun Qi.
\newblock Facial expression recognition in the wild via deep attentive center
  loss.
\newblock In {\em Proceedings of the IEEE/CVF Winter Conference on Applications
  of Computer Vision}, pages 2402--2411, 2021.

\bibitem{llhf}
Mariana-Iuliana Georgescu, Radu~Tudor Ionescu, and Marius Popescu.
\newblock Local learning with deep and handcrafted features for facial
  expression recognition.
\newblock {\em IEEE Access}, 7:64827--64836, 2019.

\bibitem{gian}
Panagiotis Giannopoulos, Isidoros Perikos, and Ioannis Hatzilygeroudis.
\newblock Deep learning approaches for facial emotion recognition: A case study
  on fer-2013.
\newblock In {\em Advances in hybridization of intelligent methods}, pages
  1--16. Springer, 2018.

\bibitem{relu}
Xavier Glorot, Antoine Bordes, and Yoshua Bengio.
\newblock Deep sparse rectifier neural networks.
\newblock In {\em Proceedings of the fourteenth international conference on
  artificial intelligence and statistics}, pages 315--323. JMLR Workshop and
  Conference Proceedings, 2011.

\bibitem{resnet}
Kaiming He, Xiangyu Zhang, Shaoqing Ren, and Jian Sun.
\newblock Deep residual learning for image recognition.
\newblock In {\em Proceedings of the IEEE conference on computer vision and
  pattern recognition}, pages 770--778, 2016.

\bibitem{senet}
Jie Hu, Li Shen, and Gang Sun.
\newblock Squeeze-and-excitation networks.
\newblock In {\em Proceedings of the IEEE conference on computer vision and
  pattern recognition}, pages 7132--7141, 2018.

\bibitem{bag}
Radu~Tudor Ionescu, Marius Popescu, and Cristian Grozea.
\newblock Local learning to improve bag of visual words model for facial
  expression recognition.
\newblock In {\em Workshop on challenges in representation learning, ICML},
  2013.

\bibitem{jung}
Heechul Jung, Sihaeng Lee, Junho Yim, Sunjeong Park, and Junmo Kim.
\newblock Joint fine-tuning in deep neural networks for facial expression
  recognition.
\newblock In {\em Proceedings of the IEEE international conference on computer
  vision}, pages 2983--2991, 2015.

\bibitem{adam}
Diederik~P Kingma and Jimmy Ba.
\newblock Adam: A method for stochastic optimization.
\newblock {\em arXiv preprint arXiv:1412.6980}, 2014.

\bibitem{alexnet}
Alex Krizhevsky, Ilya Sutskever, and Geoffrey~E Hinton.
\newblock Imagenet classification with deep convolutional neural networks.
\newblock {\em Advances in neural information processing systems},
  25:1097--1105, 2012.

\bibitem{lecun}
Yann LeCun, L{\'e}on Bottou, Yoshua Bengio, and Patrick Haffner.
\newblock Gradient-based learning applied to document recognition.
\newblock {\em Proceedings of the IEEE}, 86(11):2278--2324, 1998.

\bibitem{raf-db}
Shan Li, Weihong Deng, and JunPing Du.
\newblock Reliable crowdsourcing and deep locality-preserving learning for
  expression recognition in the wild.
\newblock In {\em Proceedings of the IEEE conference on computer vision and
  pattern recognition}, pages 2852--2861, 2017.

\bibitem{suwa}
Suwa M., Sugie N., and Fujimora K.
\newblock A preliminary note on pattern recognition of human emotional
  expression.
\newblock In {\em Proceedings of The 4th International Joint Conference on
  Pattern Recognition}, pages 408--410, 1978.

\bibitem{mase}
Kenji Mase.
\newblock Recognition of facial expression from optical flow.
\newblock {\em IEICE TRANSACTIONS on Information and Systems},
  74(10):3474--3483, 1991.

\bibitem{deep-emotion}
Shervin Minaee and Amirali Abdolrashidi.
\newblock Deep-emotion: Facial expression recognition using attentional
  convolutional network.
\newblock {\em arXiv preprint arXiv:1902.01019}, 2019.

\bibitem{molla}
Ali Mollahosseini, David Chan, and Mohammad~H Mahoor.
\newblock Going deeper in facial expression recognition using deep neural
  networks.
\newblock In {\em 2016 IEEE Winter conference on applications of computer
  vision (WACV)}, pages 1--10. IEEE, 2016.

\bibitem{affectnet}
Ali Mollahosseini, Behzad Hasani, and Mohammad~H Mahoor.
\newblock Affectnet: A database for facial expression, valence, and arousal
  computing in the wild.
\newblock {\em IEEE Transactions on Affective Computing}, 10(1):18--31, 2017.

\bibitem{sift}
Pauline~C Ng and Steven Henikoff.
\newblock Sift: Predicting amino acid changes that affect protein function.
\newblock {\em Nucleic acids research}, 31(13):3812--3814, 2003.

\bibitem{pan}
Bowen Pan, Shangfei Wang, and Bin Xia.
\newblock Occluded facial expression recognition enhanced through privileged
  information.
\newblock In {\em Proceedings of the 27th ACM International Conference on
  Multimedia}, pages 566--573, 2019.

\bibitem{pytorch}
Adam Paszke, Sam Gross, Francisco Massa, Adam Lerer, James Bradbury, Gregory
  Chanan, Trevor Killeen, Zeming Lin, Natalia Gimelshein, Luca Antiga, et~al.
\newblock Pytorch: An imperative style, high-performance deep learning library.
\newblock {\em arXiv preprint arXiv:1912.01703}, 2019.

\bibitem{lbp}
Caifeng Shan, Shaogang Gong, and Peter~W McOwan.
\newblock Facial expression recognition based on local binary patterns: A
  comprehensive study.
\newblock {\em Image and vision Computing}, 27(6):803--816, 2009.

\bibitem{vgg}
Karen Simonyan and Andrew Zisserman.
\newblock Very deep convolutional networks for large-scale image recognition.
\newblock {\em arXiv preprint arXiv:1409.1556}, 2014.

\bibitem{googlenet}
Christian Szegedy, Wei Liu, Yangqing Jia, Pierre Sermanet, Scott Reed, Dragomir
  Anguelov, Dumitru Erhan, Vincent Vanhoucke, and Andrew Rabinovich.
\newblock Going deeper with convolutions.
\newblock In {\em Proceedings of the IEEE conference on computer vision and
  pattern recognition}, pages 1--9, 2015.

\bibitem{tian}
Y-I Tian, Takeo Kanade, and Jeffrey~F Cohn.
\newblock Recognizing action units for facial expression analysis.
\newblock {\em IEEE Transactions on pattern analysis and machine intelligence},
  23(2):97--115, 2001.

\bibitem{psr}
Thanh-Hung Vo, Guee-Sang Lee, Hyung-Jeong Yang, and Soo-Hyung Kim.
\newblock Pyramid with super resolution for in-the-wild facial expression
  recognition.
\newblock {\em IEEE Access}, 8:131988--132001, 2020.

\bibitem{scn}
Kai Wang, Xiaojiang Peng, Jianfei Yang, Shijian Lu, and Yu Qiao.
\newblock Suppressing uncertainties for large-scale facial expression
  recognition.
\newblock In {\em Proceedings of the IEEE/CVF Conference on Computer Vision and
  Pattern Recognition}, pages 6897--6906, 2020.

\bibitem{ran}
Kai Wang, Xiaojiang Peng, Jianfei Yang, Debin Meng, and Yu Qiao.
\newblock Region attention networks for pose and occlusion robust facial
  expression recognition.
\newblock {\em IEEE Transactions on Image Processing}, 29:4057--4069, 2020.

\bibitem{yang}
Huiyuan Yang, Umur Ciftci, and Lijun Yin.
\newblock Facial expression recognition by de-expression residue learning.
\newblock In {\em Proceedings of the IEEE conference on computer vision and
  pattern recognition}, pages 2168--2177, 2018.

\bibitem{zeng}
Jiabei Zeng, Shiguang Shan, and Xilin Chen.
\newblock Facial expression recognition with inconsistently annotated datasets.
\newblock In {\em Proceedings of the European conference on computer vision
  (ECCV)}, pages 222--237, 2018.

\bibitem{zhangfeifei}
Feifei Zhang, Tianzhu Zhang, Qirong Mao, and Changsheng Xu.
\newblock Joint pose and expression modeling for facial expression recognition.
\newblock In {\em Proceedings of the IEEE conference on computer vision and
  pattern recognition}, pages 3359--3368, 2018.

\bibitem{nmf}
Ruicong Zhi, Markus Flierl, Qiuqi Ruan, and W~Bastiaan Kleijn.
\newblock Graph-preserving sparse nonnegative matrix factorization with
  application to facial expression recognition.
\newblock {\em IEEE Transactions on Systems, Man, and Cybernetics, Part B
  (Cybernetics)}, 41(1):38--52, 2010.

\end{thebibliography}


\begin{thebibliography}{10}\itemsep=-1pt

\bibitem{imbalance}
Mateusz Buda, Atsuto Maki, and Maciej~A Mazurowski.
\newblock A systematic study of the class imbalance problem in convolutional
  neural networks.
\newblock {\em Neural Networks}, 106:249--259, 2018.

\bibitem{island}
Jie Cai, Zibo Meng, Ahmed~Shehab Khan, Zhiyuan Li, James O'Reilly, and Yan
  Tong.
\newblock Island loss for learning discriminative features in facial expression
  recognition.
\newblock In {\em 2018 13th IEEE International Conference on Automatic Face \&
  Gesture Recognition (FG 2018)}, pages 302--309. IEEE, 2018.

\bibitem{fer2013}
P.~L. Carrier, A. Courville, I.~J. Goodfellow, M. Mirza, and Y. Bengio.
\newblock {\em FER-2013 face database.}
\newblock Universit de Montreal, 2013.

\bibitem{chenshikai}
Shikai Chen, Jianfeng Wang, Yuedong Chen, Zhongchao Shi, Xin Geng, and Yong
  Rui.
\newblock Label distribution learning on auxiliary label space graphs for
  facial expression recognition.
\newblock In {\em Proceedings of the IEEE/CVF Conference on Computer Vision and
  Pattern Recognition}, pages 13984--13993, 2020.

\bibitem{hog}
Navneet Dalal and Bill Triggs.
\newblock Histograms of oriented gradients for human detection.
\newblock In {\em 2005 IEEE computer society conference on computer vision and
  pattern recognition (CVPR'05)}, volume~1, pages 886--893. Ieee, 2005.

\bibitem{darwin}
Charles Darwin and Phillip Prodger.
\newblock {\em The expression of the emotions in man and animals}.
\newblock Oxford University Press, USA, 1998.

\bibitem{imagenet}
Jia Deng, Wei Dong, Richard Socher, Li-Jia Li, Kai Li, and Li Fei-Fei.
\newblock Imagenet: A large-scale hierarchical image database.
\newblock In {\em 2009 IEEE conference on computer vision and pattern
  recognition}, pages 248--255. Ieee, 2009.

\bibitem{sfew}
Abhinav Dhall, Roland Goecke, Simon Lucey, and Tom Gedeon.
\newblock Static facial expression analysis in tough conditions: Data,
  evaluation protocol and benchmark.
\newblock In {\em 2011 IEEE International Conference on Computer Vision
  Workshops (ICCV Workshops)}, pages 2106--2112. IEEE, 2011.

\bibitem{ekman}
Paul Ekman and Wallace~V Friesen.
\newblock Constants across cultures in the face and emotion.
\newblock {\em Journal of personality and social psychology}, 17(2):124, 1971.

\bibitem{emotionet}
C Fabian Benitez-Quiroz, Ramprakash Srinivasan, and Aleix~M Martinez.
\newblock Emotionet: An accurate, real-time algorithm for the automatic
  annotation of a million facial expressions in the wild.
\newblock In {\em Proceedings of the IEEE conference on computer vision and
  pattern recognition}, pages 5562--5570, 2016.

\bibitem{dacl}
Amir~Hossein Farzaneh and Xiaojun Qi.
\newblock Facial expression recognition in the wild via deep attentive center
  loss.
\newblock In {\em Proceedings of the IEEE/CVF Winter Conference on Applications
  of Computer Vision}, pages 2402--2411, 2021.

\bibitem{llhf}
Mariana-Iuliana Georgescu, Radu~Tudor Ionescu, and Marius Popescu.
\newblock Local learning with deep and handcrafted features for facial
  expression recognition.
\newblock {\em IEEE Access}, 7:64827--64836, 2019.

\bibitem{relu}
Xavier Glorot, Antoine Bordes, and Yoshua Bengio.
\newblock Deep sparse rectifier neural networks.
\newblock In {\em Proceedings of the fourteenth international conference on
  artificial intelligence and statistics}, pages 315--323. JMLR Workshop and
  Conference Proceedings, 2011.

\bibitem{resnet}
Kaiming He, Xiangyu Zhang, Shaoqing Ren, and Jian Sun.
\newblock Deep residual learning for image recognition.
\newblock In {\em Proceedings of the IEEE conference on computer vision and
  pattern recognition}, pages 770--778, 2016.

\bibitem{senet}
Jie Hu, Li Shen, and Gang Sun.
\newblock Squeeze-and-excitation networks.
\newblock In {\em Proceedings of the IEEE conference on computer vision and
  pattern recognition}, pages 7132--7141, 2018.

\bibitem{jung}
Heechul Jung, Sihaeng Lee, Junho Yim, Sunjeong Park, and Junmo Kim.
\newblock Joint fine-tuning in deep neural networks for facial expression
  recognition.
\newblock In {\em Proceedings of the IEEE international conference on computer
  vision}, pages 2983--2991, 2015.

\bibitem{adam}
Diederik~P Kingma and Jimmy Ba.
\newblock Adam: A method for stochastic optimization.
\newblock {\em arXiv preprint arXiv:1412.6980}, 2014.

\bibitem{afew}
Jean Kossaifi, Georgios Tzimiropoulos, Sinisa Todorovic, and Maja Pantic.
\newblock Afew-va database for valence and arousal estimation in-the-wild.
\newblock {\em Image and Vision Computing}, 65:23--36, 2017.

\bibitem{alexnet}
Alex Krizhevsky, Ilya Sutskever, and Geoffrey~E Hinton.
\newblock Imagenet classification with deep convolutional neural networks.
\newblock {\em Advances in neural information processing systems},
  25:1097--1105, 2012.

\bibitem{lecun}
Yann LeCun, L{\'e}on Bottou, Yoshua Bengio, and Patrick Haffner.
\newblock Gradient-based learning applied to document recognition.
\newblock {\em Proceedings of the IEEE}, 86(11):2278--2324, 1998.

\bibitem{raf-db}
Shan Li, Weihong Deng, and JunPing Du.
\newblock Reliable crowdsourcing and deep locality-preserving learning for
  expression recognition in the wild.
\newblock In {\em Proceedings of the IEEE conference on computer vision and
  pattern recognition}, pages 2852--2861, 2017.

\bibitem{suwa}
Suwa M., Sugie N., and Fujimora K.
\newblock A preliminary note on pattern recognition of human emotional
  expression.
\newblock In {\em Proceedings of The 4th International Joint Conference on
  Pattern Recognition}, pages 408--410, 1978.

\bibitem{mase}
Kenji Mase.
\newblock Recognition of facial expression from optical flow.
\newblock {\em IEICE TRANSACTIONS on Information and Systems},
  74(10):3474--3483, 1991.

\bibitem{identity}
Zibo Meng, Ping Liu, Jie Cai, Shizhong Han, and Yan Tong.
\newblock Identity-aware convolutional neural network for facial expression
  recognition.
\newblock In {\em 2017 12th IEEE International Conference on Automatic Face \&
  Gesture Recognition (FG 2017)}, pages 558--565. IEEE, 2017.

\bibitem{affectnet}
Ali Mollahosseini, Behzad Hasani, and Mohammad~H Mahoor.
\newblock Affectnet: A database for facial expression, valence, and arousal
  computing in the wild.
\newblock {\em IEEE Transactions on Affective Computing}, 10(1):18--31, 2017.

\bibitem{sift}
Pauline~C Ng and Steven Henikoff.
\newblock Sift: Predicting amino acid changes that affect protein function.
\newblock {\em Nucleic acids research}, 31(13):3812--3814, 2003.

\bibitem{pan}
Bowen Pan, Shangfei Wang, and Bin Xia.
\newblock Occluded facial expression recognition enhanced through privileged
  information.
\newblock In {\em Proceedings of the 27th ACM International Conference on
  Multimedia}, pages 566--573, 2019.

\bibitem{pytorch}
Adam Paszke, Sam Gross, Francisco Massa, Adam Lerer, James Bradbury, Gregory
  Chanan, Trevor Killeen, Zeming Lin, Natalia Gimelshein, Luca Antiga, et~al.
\newblock Pytorch: An imperative style, high-performance deep learning library.
\newblock {\em arXiv preprint arXiv:1912.01703}, 2019.

\bibitem{lbp}
Caifeng Shan, Shaogang Gong, and Peter~W McOwan.
\newblock Facial expression recognition based on local binary patterns: A
  comprehensive study.
\newblock {\em Image and vision Computing}, 27(6):803--816, 2009.

\bibitem{sub}
Wenzhe Shi, Jose Caballero, Ferenc Husz{\'a}r, Johannes Totz, Andrew~P Aitken,
  Rob Bishop, Daniel Rueckert, and Zehan Wang.
\newblock Real-time single image and video super-resolution using an efficient
  sub-pixel convolutional neural network.
\newblock In {\em Proceedings of the IEEE conference on computer vision and
  pattern recognition}, pages 1874--1883, 2016.

\bibitem{vgg}
Karen Simonyan and Andrew Zisserman.
\newblock Very deep convolutional networks for large-scale image recognition.
\newblock {\em arXiv preprint arXiv:1409.1556}, 2014.

\bibitem{googlenet}
Christian Szegedy, Wei Liu, Yangqing Jia, Pierre Sermanet, Scott Reed, Dragomir
  Anguelov, Dumitru Erhan, Vincent Vanhoucke, and Andrew Rabinovich.
\newblock Going deeper with convolutions.
\newblock In {\em Proceedings of the IEEE conference on computer vision and
  pattern recognition}, pages 1--9, 2015.

\bibitem{tian}
Y-I Tian, Takeo Kanade, and Jeffrey~F Cohn.
\newblock Recognizing action units for facial expression analysis.
\newblock {\em IEEE Transactions on pattern analysis and machine intelligence},
  23(2):97--115, 2001.

\bibitem{psr}
Thanh-Hung Vo, Guee-Sang Lee, Hyung-Jeong Yang, and Soo-Hyung Kim.
\newblock Pyramid with super resolution for in-the-wild facial expression
  recognition.
\newblock {\em IEEE Access}, 8:131988--132001, 2020.

\bibitem{scn}
Kai Wang, Xiaojiang Peng, Jianfei Yang, Shijian Lu, and Yu Qiao.
\newblock Suppressing uncertainties for large-scale facial expression
  recognition.
\newblock In {\em Proceedings of the IEEE/CVF Conference on Computer Vision and
  Pattern Recognition}, pages 6897--6906, 2020.

\bibitem{ran}
Kai Wang, Xiaojiang Peng, Jianfei Yang, Debin Meng, and Yu Qiao.
\newblock Region attention networks for pose and occlusion robust facial
  expression recognition.
\newblock {\em IEEE Transactions on Image Processing}, 29:4057--4069, 2020.

\bibitem{yang}
Huiyuan Yang, Umur Ciftci, and Lijun Yin.
\newblock Facial expression recognition by de-expression residue learning.
\newblock In {\em Proceedings of the IEEE conference on computer vision and
  pattern recognition}, pages 2168--2177, 2018.

\bibitem{yang2018}
Huiyuan Yang, Umur Ciftci, and Lijun Yin.
\newblock Facial expression recognition by de-expression residue learning.
\newblock In {\em Proceedings of the IEEE conference on computer vision and
  pattern recognition}, pages 2168--2177, 2018.

\bibitem{multiple}
Zhiding Yu and Cha Zhang.
\newblock Image based static facial expression recognition with multiple deep
  network learning.
\newblock In {\em Proceedings of the 2015 ACM on international conference on
  multimodal interaction}, pages 435--442, 2015.

\bibitem{zeng}
Jiabei Zeng, Shiguang Shan, and Xilin Chen.
\newblock Facial expression recognition with inconsistently annotated datasets.
\newblock In {\em Proceedings of the European conference on computer vision
  (ECCV)}, pages 222--237, 2018.

\bibitem{zhangfeifei}
Feifei Zhang, Tianzhu Zhang, Qirong Mao, and Changsheng Xu.
\newblock Joint pose and expression modeling for facial expression recognition.
\newblock In {\em Proceedings of the IEEE conference on computer vision and
  pattern recognition}, pages 3359--3368, 2018.

\bibitem{nmf}
Ruicong Zhi, Markus Flierl, Qiuqi Ruan, and W~Bastiaan Kleijn.
\newblock Graph-preserving sparse nonnegative matrix factorization with
  application to facial expression recognition.
\newblock {\em IEEE Transactions on Systems, Man, and Cybernetics, Part B
  (Cybernetics)}, 41(1):38--52, 2010.

\end{thebibliography}
}

\end{document}